\documentclass{article}

 \usepackage[final,nonatbib]{metalearn_2019}

\usepackage[utf8]{inputenc} % allow utf-8 input
\usepackage[T1]{fontenc}    % use 8-bit T1 fonts
\usepackage{hyperref}       % hyperlinks
\usepackage{url}            % simple URL typesetting
\usepackage{booktabs}       % professional-quality tables
\usepackage{amsfonts}       % blackboard math symbols
\usepackage{nicefrac}       % compact symbols for 1/2, etc.
\usepackage{microtype}      % microtypography
\usepackage{graphicx}
\usepackage{subfigure}
\usepackage{algorithm}
\usepackage{algorithmicx}
\usepackage{algpseudocode}

\usepackage{multirow}

\title{Learning to Tune XGBoost with XGBoost}

\author{
Johanna Sommer$^{1,2}$, Dimitrios Sarigiannis$^3$, Thomas Parnell$^3$\\
   $^1$IBM Germany,
   $^2$Duale Hochschule Baden-Württemberg,
   $^3$IBM Research\\
    \texttt{johanna@mail-sommer.com},   
	\texttt{saridimi@gmail.com},
    \texttt{tpa@zurich.ibm.com}\\ 
}

\begin{document}

\maketitle

\begin{abstract}
In this short paper we investigate whether meta-learning techniques can be used to more effectively tune the hyperparameters of machine learning models using successive halving (SH).
We propose a novel variant of the SH algorithm (MeSH), that uses meta-regressors to determine which candidate configurations should be eliminated at each round.
We apply MeSH to the problem of tuning the hyperparameters of a gradient-boosted decision tree model.
By training and tuning our meta-regressors using existing tuning jobs from 95 datasets, we demonstrate that MeSH can often find a superior solution to both SH and random search.
\end{abstract}

\section{Introduction}

Modern machine learning models have a large number of hyperparameters that must be tuned. 
Gradient-boosted decision trees (GBDTs), perhaps the most popular classical machine learning model today, have at least five hyperparameters that must be tuned. 
Similarly, one can view the architecture of a deep neural network as a highly-structured hyperparameter space. 
This insight has led to the emerging field of neural architecture search, in which researchers attempt to let optimization algorithms discover neural architectures that out-perform those designed by their colleagues. 

\begin{figure}
  \centering
  \subfigure[Ideal behaviour: the best configuration in the final round can be identified from the beginning and is not eliminated as the algorithm progresses.] {
  	\includegraphics[width=0.46\columnwidth]{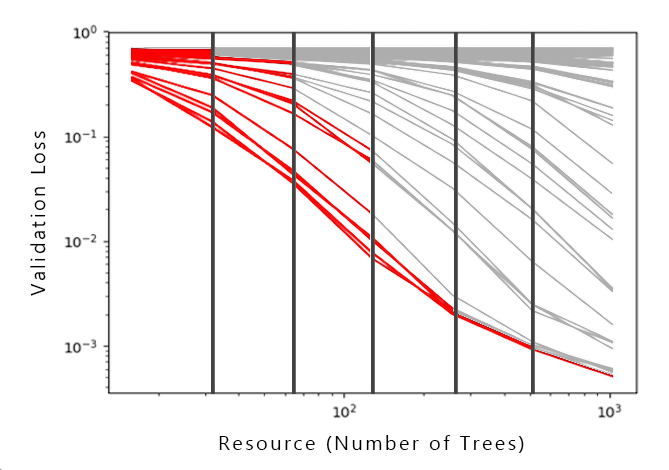}
	\label{fig:sh_good}
  }
  \hskip 0.5cm
  \subfigure[Problematic behaviour: promising configurations in the initial rounds do not correspond to the best configurations in the final round.] {
  	\includegraphics[width=0.46\columnwidth]{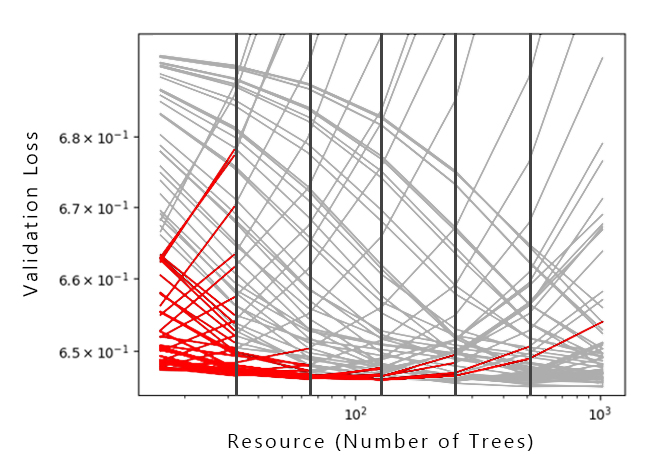}
	\label{fig:sh_bad}
  }
  \caption{Example of how configurations are eliminated in the SH algorithm, applied to tuning XGBoost \cite{Chen:2016:XST:2939672.2939785}, for two different datasets. Vertical lines indicate the rounds at which half of the configurations are discarded, configurations that survive elimination are marked in red.}
  \label{fig:sh_good_bad}
  %\vspace{-5mm}
\end{figure}

Model-free methods for hyperparameter optimization based on successive halving (SH) \cite{DBLP:journals/corr/LiJDRT16, DBLP:journals/corr/abs-1810-05934} have recently gained a lot of traction in the aforementioned applications. 
Their popularity can be attributed to the ease with which they can be parallelized and/or distributed across a cluster of machines, as well as the lack of assumptions made about the nature of the underlying space relative to model-based approaches such as Bayesian optimization \cite{HutHooLey11}.
Essential to these methods is the notion of a resource \cite{DBLP:journals/corr/LiJDRT16}: normally either the number of iterations of the learning algorithm or the number of training examples.
By randomly sampling a large number of hyperparameter configurations and evaluating them on a small resource, one can identify promising configurations quickly, and carry forward only the best for evaluation on a larger resource. 
Implicit to this algorithm is the idea that the relative ordering of configurations (in terms of validation loss) is consistent whether using a small or a large resource. 
However, when tuning GBDTs we often observe that this is far from reality, as illustrated in Figure \ref{fig:sh_good_bad}
.

Meta-learning \cite{DBLP:journals/corr/abs-1810-03548} is an orthogonal field that focuses on \textit{learning to learn}, or learning from experience.
In the context of hyperparameter optimization this means: can we leverage our experience from previous tuning tasks on different datasets, to better tune the hyperparameters of a given model on a new, target dataset?
Existing attempts to achieve this generally involve trying to warm-start hyperparameter optimization methods with a set of configurations that were known to perform well on \textit{similar} datasets \cite{Feurer:2015:IBH:2887007.2887164, Thornton:2013:ACS:2487575.2487629, NIPS2015_5872}. 
The task of identifying similar datasets is normally achieved by computing meta-features that characterize each dataset in an expressive way.
Of particular interest in this paper is a class of meta-features, known as \textit{landmarking meta-features} \cite{Pfahringer:2000:MLV:645529.658105, bensusan2000discovering}.
Computing a landmarking meta-feature typically involves training a relatively inexpensive model (e.g. naive Bayes) and evaluating its performance on some hold-out data. 
Such meta-features are known to play a significant role in predicting the performance of an algorithm \cite{Kalousis2002AlgorithmSV}.
However, these meta-features are relatively expensive to compute, especially if one would like to generate landmarking meta-features using more complex models.

The critical insight behind this work is as follows: in the context of hyperparameter optimization using SH, such meta-features are essentially available at no cost. 
The SH algorithm is fundamentally based on evaluating configurations with an ever-increasing resource, and all previous evaluations can be regarded as landmarking meta-features.
In the following, we explore how we may exploit this observation, to allow SH to learn from experience.

\section{Improving Successive Halving via Meta-Learning}

In this section we describe a new algorithm, MeSH, that aims to use \textbf{Me}ta-learning to improve the behaviour of \textbf{S}uccessive \textbf{H}alving.
The algorithm consists of an \textit{online phase}, essentially a minor modification to the SH algorithm in which the decision regarding which configurations should be eliminated is based on the output of a set of predictive meta-models, and an \textit{offline phase} during which the meta-models are trained and tuned.
We will now describe the two phases of MeSH in more detail.

\paragraph{Online phase.}
The online phase of MeSH is presented in full in Algorithm \ref{alg:mesh}. 
If we compare and contrast the algorithm with that proposed in \cite{inproceedings, pmlr-v51-jamieson16, DBLP:journals/corr/LiJDRT16, DBLP:journals/corr/abs-1810-05934}, we will notice two key differences.
Firstly, we must provide a meta-model for each \textit{round} of the algorithm.
Secondly, when we eliminate configurations, instead of using the validation loss evaluated with the current resource, we use the output of the corresponding meta-model.
For each configuration, this meta-model takes as an input a meta-feature vector consisting of (a) some dataset meta-features, (b) configuration meta-features (i.e., the setting of each hyperparameter) and (c) the validation loss for this configuration evaluated at the previous rounds. 
The meta-model then outputs, for each configuration, a prediction of how this configuration would perform (in terms of validation loss) if evaluated using the maximal resource.
These predicted values are then sorted and only the top $1/\eta$ are carried over into the next round.

\begin{algorithm}[ht]
\begin{algorithmic}[1]
\Require{Number of configurations $n$, eliminator factor $\eta$, min. resource $r_{min}$, max. resource $r_{max}$}
\State $s_{max} \gets \lfloor \log_{\eta}(r_{max} / r_{min})\rfloor$
\Require{Meta-models $M_i$ for $i=0,1\ldots,s_{max}$, Dataset meta-features $\mathcal{D}$}
\Ensure{$n\geq \eta^{s_{max}}$}
\State $T \gets sample\_configurations(n)$
\For{$i \in \{0,1,...,s_{max}\}$}                    
\State $n_i \gets \lfloor n \eta^{-i}\rfloor$
\State $r_i \gets r_{min}\eta^{i}$
\State $L_i \gets eval\_and\_return\_val\_loss(\theta, r_i):\theta \in T$
\State $L_i^\prime \gets predict\_final\_val\_loss(M_i; \mathcal{D}, T, L_{i-1}, L_{i-2},\ldots,L_0)$
\State $T \gets top\_k(T,L_i^\prime,n_i/\eta)$
\EndFor
\State \Return {Configuration with the lowest validation loss seen throughout the course of the algorithm.}
\end{algorithmic}
\caption{MeSH (Online Phase)}
\label{alg:mesh}
\end{algorithm}

\begin{figure}
	\centering
	\includegraphics[width=0.8\textwidth]{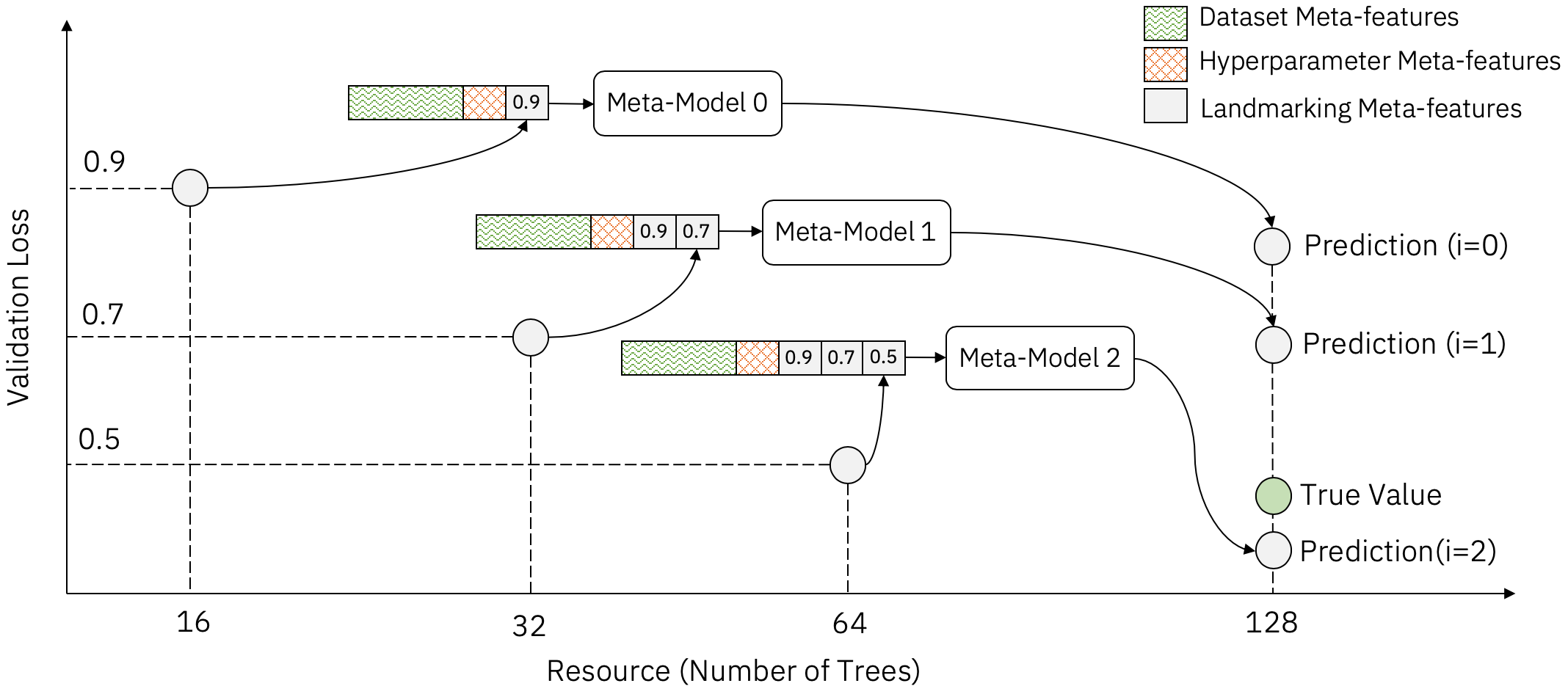}
    \caption{MeSH uses meta-regressors to predict the final validation loss.}
	\label{fig:mesh_online} 
	\vspace{-6mm}
\end{figure}

The online phase is graphically illustrated in Figure \ref{fig:mesh_online}, for the case where $r_{min}=16$, $r_{max}=128$ and $\eta=2$. 
The general idea is that as more and more landmarking features become available (for the later rounds), the more accurate the predictions of the meta-models.
It is hoped that, by using the predicted value of the validation loss at the maximal resource, rather than using the measured validation loss at the current resource, that we can overcome the \textit{crossover} problem that was illustrated in Figure \ref{fig:sh_bad}.
In terms of overheads relative to vanilla SH, since the landmarking meta-features are given for free, one must only pay the price of making predictions with the meta-model, which in general is negligible compared to the cost of training.

\paragraph{Offline phase.}
The approach above can only prove useful if the meta-models are indeed accurate at predicting the final-round validation loss.
In order to achieve this, in the offline phase of MeSH, we train and tune the meta-models using a large amount of data.
Firstly, in order to construct the required meta-datasets, one must conduct a large experiment.
This experiment involves sampling hyperparameter configurations, uniformly at random, for a large number of datasets and evaluating their validation loss given resource $r_i$ for $i=0,1,\ldots,s_{\max}$.
For the $i$-th meta-model, once then constructs a meta-dataset for which the meta-feature vectors consists of (a) dataset meta-features, (b) hyperparameter configuration meta-features and (c) all validation losses evaluated with resource $r_j$ for $j=0,1\ldots,(i-1)$. 
The meta-model is then trained to minimize the mean-squared error between the predicted values and the target, given by the validation loss for the corresponding configuration evaluated at $r_{max}$. 
%Similar approaches to predicting performance using meta-regressors were previously studied in \cite{portfolio_sat}.
In order to tune the hyperparameters of the meta-model itself, one can use standard cross-validation techniques.

\section{Experimental Results}

\paragraph{Hyperparameter tuning for XGBoost.}
In this section we consider the problem of tuning the hyperparameters of an XGBoost model. 
Namely, we wish to tune: \texttt{lambda}, \texttt{colsample\_bytree}, \texttt{max\_depth} and \texttt{learning\_rate} and \texttt{num\_boost\_rounds}.
We will compare three solutions: random search (RS), SH and MeSH.
We take \texttt{num\_boost\_rounds} to be the resource (as defined in SH and MeSH), and compare schemes under the constraint that the total resource utilized (budget) is the same.
For SH and MeSH, we take $n=64$, $\eta=2$, $r_{min}=16$ and $r_{max}=1024$, resulting in 7 rounds in both cases.
For RS, the scheme of equivalent budget corresponds to 7 configurations evaluated using the maximal resource\footnote{We leverage the early-stopping functionality of XGBoost so that the training can stop early (i.e., for a smaller number of boosting rounds) if the validation score starts to increase.}.
In all cases we use logistic loss as the training and validation loss function. 

\paragraph{Constructing the meta-datasets.}
In order to generate robust meta-datasets we ran a large experiment using 95 binary classification datasets from the OpenML platform \cite{OpenML2013}.
For each dataset, we ran an experiment for 40 CPU-hours, randomly sampling from the hyperparameter space, and recording the validation loss when trained with resource $r_i=16,32,\ldots,1024$.
We then create the meta-feature vectors for the meta-dataset, comprising the landmarking meta-features extracted from the aforementioned experiment, together with the hyperparameter meta-features and the meta-features that describe the dataset. For the latter, we use a subset of the meta-features suggested in \cite{DBLP:journals/corr/abs-1810-03548}.
The resulting meta-dataset has around 300 thousand examples and up to 26 meta-features.

\begin{figure}
  \centering
  \subfigure[Offline phase: meta-models become more accurate with more landmarking meta-features.] {
  	\includegraphics[width=0.45\columnwidth]{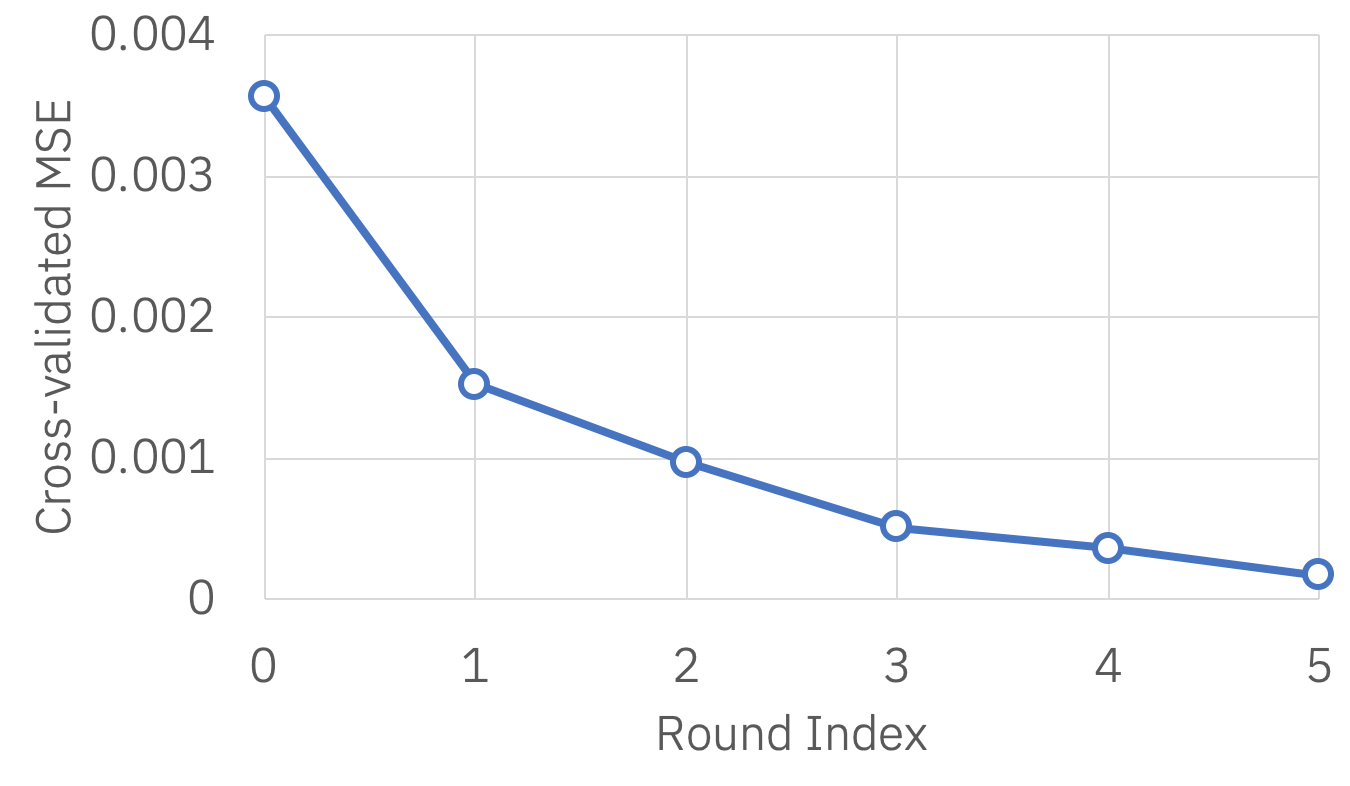}
	\label{fig:perf_offline}
  }
  \hskip 0.5cm
  \subfigure[Online phase: validation loss achieved by RS, SH and MeSH on Dataset 904 (10 repetitions).] {
  	\includegraphics[width=0.45\columnwidth]{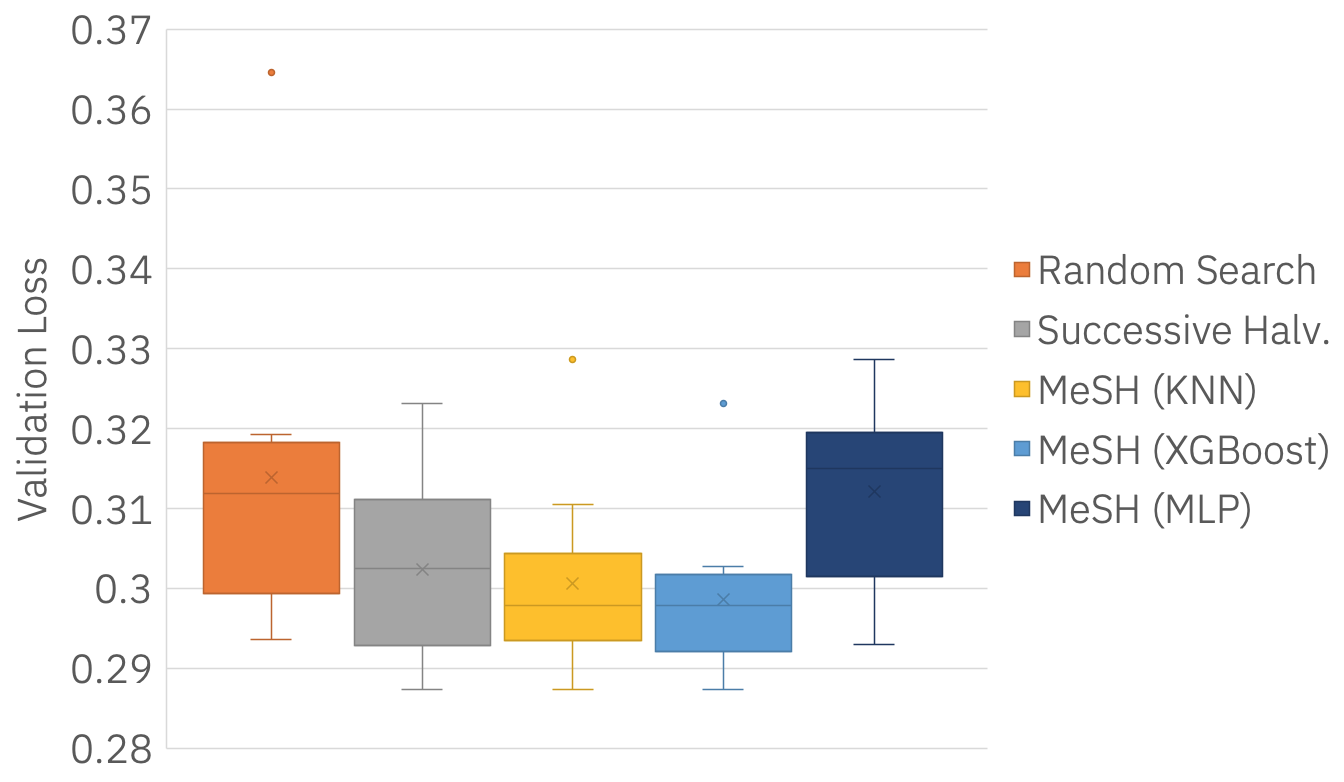}
	\label{fig:perf_online}
  }
  \caption{Performance of MeSH during offline phase and online phase.}
  \label{fig:perf}
\end{figure}

\paragraph{Training and tuning the meta-models.}
We consider three candidate meta-regressors: K-nearest neighbors (KNN), a multi-layer perceptron (MLP) and XGBoost itself.
When training and tuning the meta-models, it is critical that the datasets that we wish evaluate in the online phase are not contained within the meta-dataset.
We thus identify 5 test datasets.
For each test dataset, we remove all of the corresponding examples from the meta-dataset, and perform 3-fold cross-validation on the remaining examples to tune the hyperparameters of the meta-regressors.\footnote{Due to time restrictions, for the MLP meta-regressor we did not perform cross-validation and simply trained the meta-model with the default parameters.}
In Figure \ref{fig:perf_offline} we plot the validation MSE obtained during the offline tuning phase as a function of the round index. 
As expected, we see that as more landmarking meta-features become available, the meta-regressors are able to predict the validation loss at 1024 boosting rounds very effectively.

%In order to create robust meta-learning models, we use 95 binary classification datasets from the OpenML platform.
%By randomly sampling all the hyperparameters that we want to tune, we run XGBoost on all the num\_round values for the successive halving schedule for 40 hours per dataset and we store the %performance of each task into a database (NUMBER experiments in total). We then extract from the data (NumberK instances) the landmarking meta-features and the log-loss on the full resource, and by computing on the fly the dataset meta-features and the task meta-features we construct the full meta-dataset.
%In order to get the maximum performance from each of the meta-learners, we need to make sure that we use their optimal set of hyperparameters when train them on the meta-datasets.
%We always leave one dataset out for testing and then we split the rest of the dataset ids into two sub-sets (train and validation). We then create two meta-datasets by ensuring that no dataset exists on both. We train our meta-learners on the training meta-dataset and we evaluate their performance on the validation meta-dataset by computing the MSE score (even though our initial goal is to tune XGBoost models for binary classification tasks, our meta-learners have to be optimized for a regression task). We repeat the same process by applying random-search for a limited time.

\renewcommand{\tabcolsep}{3pt}
\begin{table}
	\centering
	\small
\begin{tabular}{|l|cc|cc|cc|cc|cc|}
\hline
\multirow{2}{*}{\textbf{Tuning Algorithm}} & \multicolumn{2}{c|}{\textbf{Dataset 799}} & \multicolumn{2}{c|}{\textbf{Dataset 904}} & \multicolumn{2}{c|}{\textbf{Dataset 813}} & \multicolumn{2}{c|}{\textbf{Dataset 897}} & \multicolumn{2}{c|}{\textbf{Dataset 930}} \\
                                           & \textbf{Mean}        & \textbf{Std}       & \textbf{Mean}        & \textbf{Std}       & \textbf{Mean}        & \textbf{Std}       & \textbf{Mean}        & \textbf{Std}       & \textbf{Mean}        & \textbf{Std}       \\
\hline
Random Search                                & 0.2209               & 0.015              & 0.3139               & 0.020              & 0.2073               & 0.015              & 0.0295               & 0.003              & 0.4903               & 0.005              \\
Successive Halving                                & \textbf{0.2140}               & 0.008              & 0.3024               & 0.011              & 0.1978               & 0.011              & \textbf{0.0262}               & 0.002              & 0.4907               & 0.004              \\
MeSH (KNN)                        & 0.2171               & 0.007              & 0.3007               & 0.012              & 0.2034               & 0.013              & 0.0264               & 0.003              & 0.4939               & 0.006              \\
MeSH (MLP)                        & 0.2173               & 0.008              & 0.3122               & 0.010              & 0.2076               & 0.010              & 0.0264               & 0.002              & 0.4901               & 0.004              \\
MeSH (XGBoost)                    & \textbf{0.2140}               & 0.008              & \textbf{0.2986}               & 0.011              & \textbf{0.1975}               & 0.010              & \textbf{0.0262}               & 0.002              & \textbf{0.4881}               & 0.004             \\ \hline
\end{tabular}
\vspace{1mm}
\caption{Full validation loss results for the 5 test datasets (10 repetitions).}
\label{tab:res}
\vspace{-8mm}
\end{table}

\paragraph{Evaluation of the online phase.}
In Figure \ref{fig:perf_online} we present the best validation loss found for one of the test datasets. Our findings are twofold: (1) MeSH is able to find a significantly better solution than both RS and SH, (2) XGBoost is the best choice of meta-regressor. In Table \ref{tab:res}, we present the full set of results for all 5 test datasets. We observe that: (a) in 3 of the 5 cases, MeSH with XGBoost is able to outperform both RS and SH, and (b) in the 2 remaining cases, MeSH is no worse than SH.

\section{Conclusion and Future Work}
We have proposed a novel modification of the SH algorithm (MeSH) that uses meta-regressors, trained and tuned on previous hyperparameter tuning jobs, in order to more effectively determine which configurations should be eliminated.
We have applied this new algorithm to the problem of tuning the hyperparameters of XGBoost and have shown it is often capable of finding a better solution than both SH and RS.
As a next step, we would like to see how MeSH performs when applied to more challenging problems like neural architecture search.

\newpage
\bibliographystyle{plain}
\bibliography{main}

\end{document}